\documentclass{article} 
\usepackage{iclr2021_conference,times}

\usepackage{cite}
\usepackage{amsmath,amssymb,amsfonts}
\usepackage{algorithmic}
\usepackage{graphicx}
\usepackage{textcomp}
\usepackage{xcolor}

\usepackage[caption=false]{subfig}
\usepackage{balance}  
\usepackage[ruled,vlined,linesnumbered]{algorithm2e}
\usepackage{multirow}
\usepackage{hhline}

\usepackage[font=small,skip=4pt]{caption}

\newcommand{\testevent}{Indy500-2019}
\newcommand{\testcar}{12}
\usepackage{colortbl} 
\definecolor{Gray}{gray}{0.9}

\usepackage{amsmath,amsfonts,bm}

\def\eqref#1{equation~\ref{#1}}

\def\1{\bm{1}}

\DeclareMathAlphabet{\mathsfit}{\encodingdefault}{\sfdefault}{m}{sl}
\SetMathAlphabet{\mathsfit}{bold}{\encodingdefault}{\sfdefault}{bx}{n}

\usepackage{hyperref}
\usepackage{url}

\title{Rank Position Forecasting in Car Racing}

\author{Antiquus S.~Hippocampus, Natalia Cerebro \& Amelie P. Amygdale \thanks{ Use footnote for providing further information
about author (webpage, alternative address)---\emph{not} for acknowledging
funding agencies.  Funding acknowledgements go at the end of the paper.} \\
Department of Computer Science\\
Cranberry-Lemon University\\
Pittsburgh, PA 15213, USA \\
\texttt{\{hippo,brain,jen\}@cs.cranberry-lemon.edu} \\
\And
Ji Q. Ren \& Yevgeny LeNet \\
Department of Computational Neuroscience \\
University of the Witwatersrand \\
Joburg, South Africa \\
\texttt{\{robot,net\}@wits.ac.za} \\
\AND
Coauthor \\
Affiliation \\
Address \\
\texttt{email}
}

\begin{document}

\maketitle

\begin{abstract}
Forecasting is challenging since uncertainty resulted from exogenous factors exists. In this work, we investigate the rank position forecasting problem in car racing, which predicts the rank positions at the future laps for cars. Among the many factors that bring changes to the rank positions, pit stops are critical but are irregular and rare events. We found existing methods, including statistical models, machine learning regression models, and state-of-the-art deep forecasting model based on encoder-decoder architecture, all have limitations in the forecasting. By elaborative analysis of pit stops events, we propose a deep model, RankNet, with the cause effects decomposition that modeling the rank position sequence and pit stop events separately.
It also incorporates probabilistic forecasting to model the uncertainty inside each sub-model. 
Through extensive experiments, RankNet demonstrates a strong performance improvement over the baselines, e.g., MAE improves more than 10\% consistently, and is also more stable when adapting to unseen new data. Details of model optimization, performance profiling are presented. It is promising not only to provide useful forecasting tools for the car racing analysis but also to shine a light on solutions to similar challenging issues in general forecasting problems.
\end{abstract}

\section{Introduction}
Deep learning based forecasting has been successfully applied in different domains, such as
demand prediction \citet{laptev_time-series_2017}, traffic prediction \citet{liao_deep_2018}, clinical state progression prediction\citet{ryan_ctc-attention_2019}, epidemic forecasting \citet{adhikari_epideep:_2019}, etc.
However, when addressing the forecasting problem in the specific domain of motorsports, we found that the state-of-the-art models in this field are based on simulation methods or machine learning methods, all highly depend on the domain knowledge\citet{heilmeier_race_2018, bekker_planning_2009,Phillips_Simulator_2014, tulabandhula_tire_2014}.
Simply applying a deep learning model here does not deliver better forecasting performance.
The forecasting problem in motorsports is challenging. 
First, the status of a race is highly dynamic, which is the collective effect of many factors, including the skills of the drivers, the configuration of the cars, the interaction among the racing cars, the dynamics of the racing strategies and events out of control, such as mechanical failures and unfortunate crashes that are hardly avoidable during the high-speed racing. 
\textbf{Uncertainty} resulted from exogenous factors is a critical obstacle for the model to forecast the future accurately. A successful model needs to incorporate these cause effects and express uncertainty. 
Secondly, motorsports forecasting is a sparse data problem, that available data are limited because, in each race, only one trajectory for each car can be observed. Moreover, some factors, such as pit stop and crash, make huge impacts on the race dynamic but are irregular and rare, which appear less than 5\% in available data. Modeling these \textbf{extreme events} are a critical part of a model.
In this work, taking IndyCar series\citet{IndyCar_Understanding_nodate} as a use case, we investigate these challenging issues in the forecasting problem. 
Our main contributions are summarized as follows:
\begin{itemize}
    \setlength{\itemsep}{0pt}
    \setlength{\parskip}{0pt}
    \item build car racing forecasting models based on deep encoder-decoder network architecture. 
    \item explore the solutions of deep learning models by model decomposition according to the cause effects relationship and modeling the uncertainty rooted in the dynamics of the racing.
    \item improve forecasting performance significantly compared with statistical, machine learning and deep learning baselines with MAE improvements more than 10\%.
    \item be promising not only to provide useful forecasting tools for the car racing industry, but also shine a light on solutions to similar challenging issues in general forecasting problems.
\end{itemize}    
\section{Problem Statement}
\subsection{Background}
Indy500 is the premier event of the IndyCar series. Each year, 33 cars compete on a 2.5-mile oval track for 200 laps. 
The track is split into several sections, or timeline. E.g., SF/SFP indicate the start and finish line on the track or on the pit lane respectively.
A local communication network broadcasts race information to all the teams, following a general data exchange protocol\citet{IndyCar_Understanding_nodate}.
\begin{figure*}[ht]
\centering
\subfloat[Rank can be calculated by LapTime and TimeBehindLeader. LapStatus and TrackStatus indicate racing status of pitstops and cuation laps.]{\includegraphics[width=0.33\linewidth]{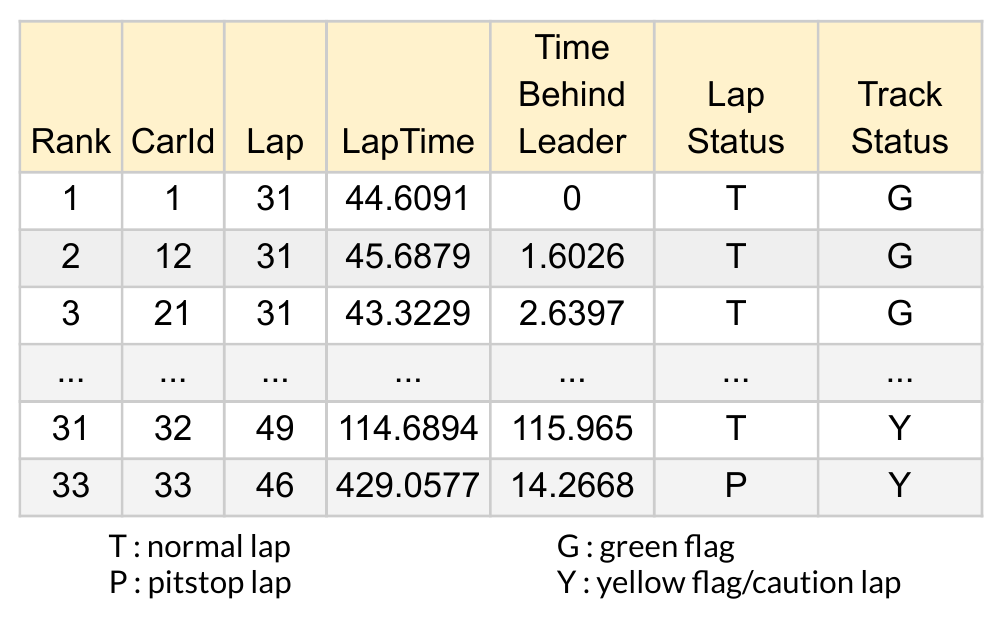}}
\hspace{2ex}
\subfloat[Rank and LapTime sequence of car12, the final winner. Sequence dynamics correlate to racing status.]{\includegraphics[width=0.51\linewidth]{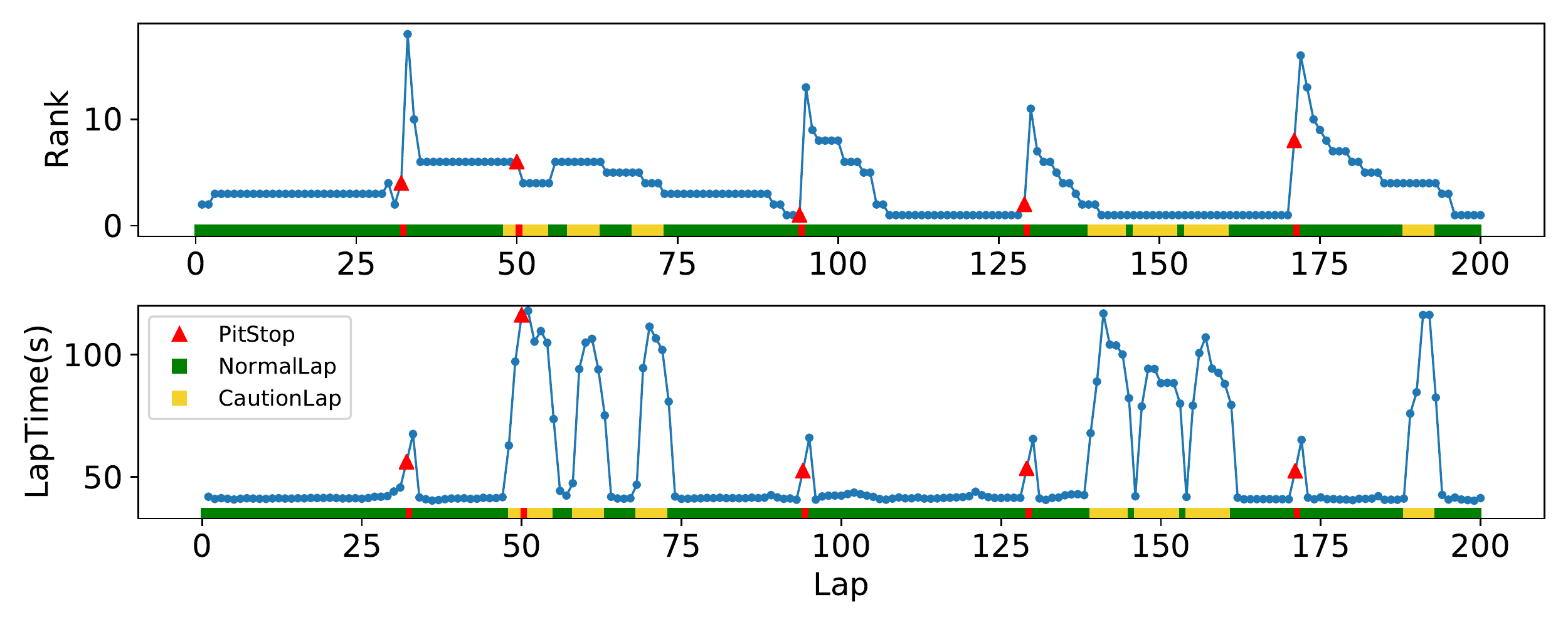}}
\caption{Data examples of Indy500-2018. }
\label{fig:indycar_data}
\end{figure*}
\begin{figure*}[ht]
\centering
 \includegraphics[width=0.24\linewidth]{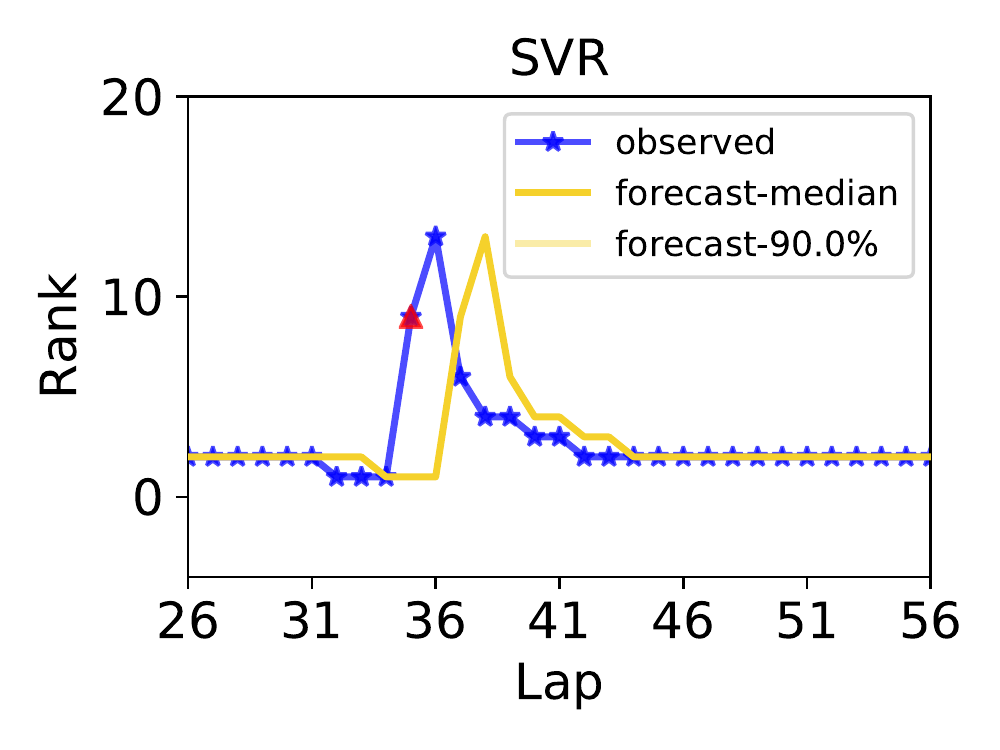}  
 \includegraphics[width=0.24\linewidth]{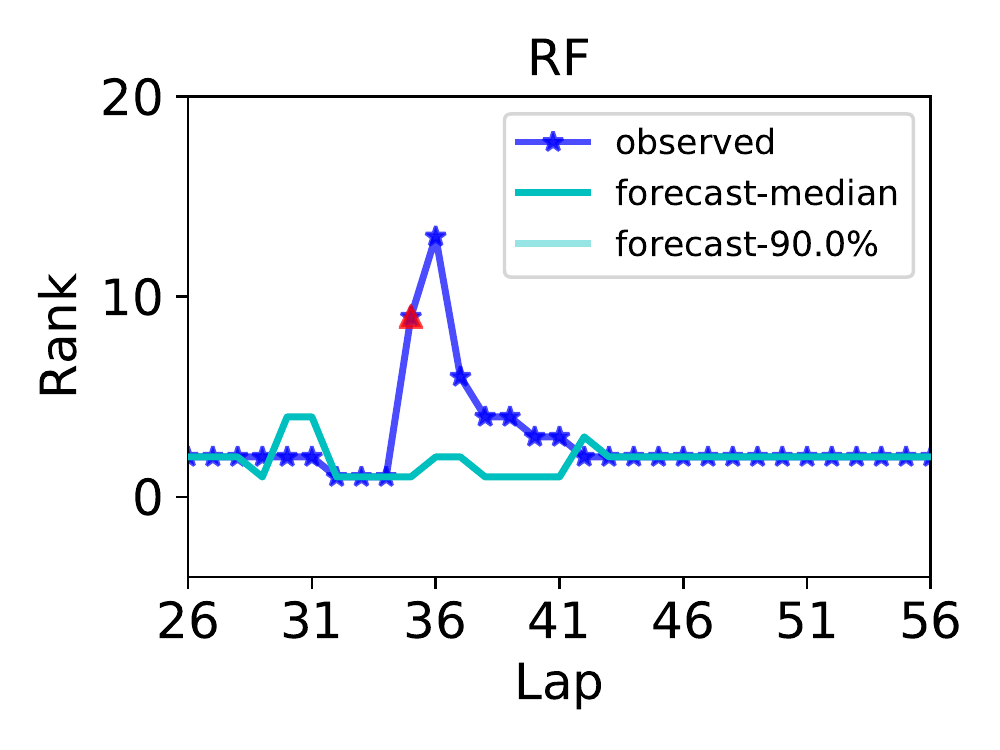}  
 \includegraphics[width=0.24\linewidth]{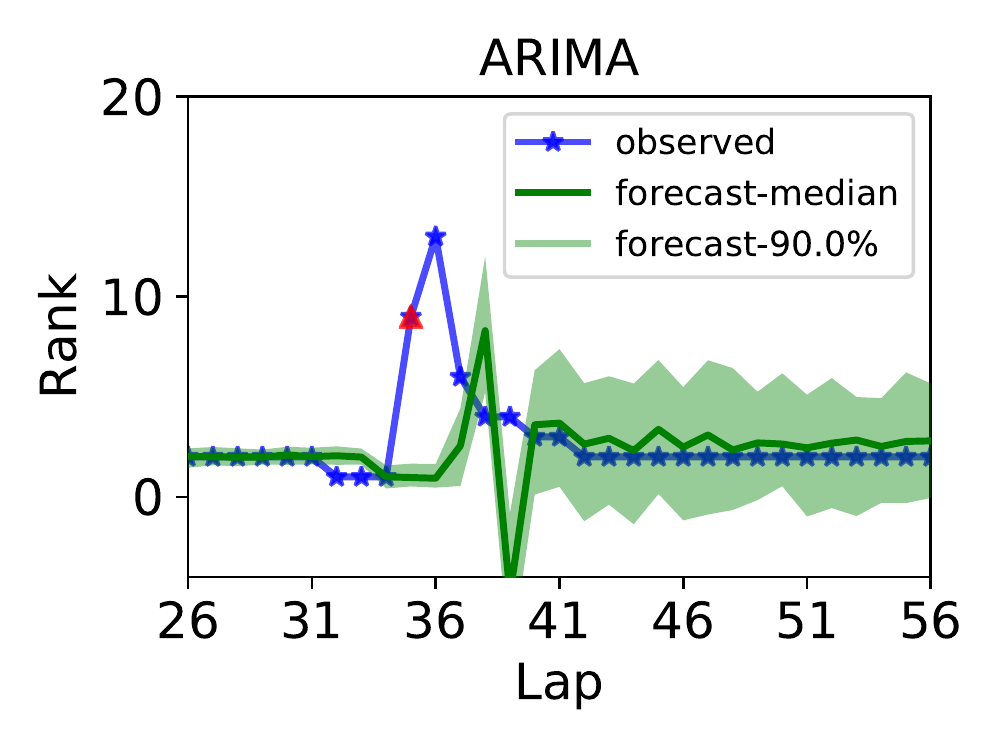}  
 \includegraphics[width=0.24\linewidth]{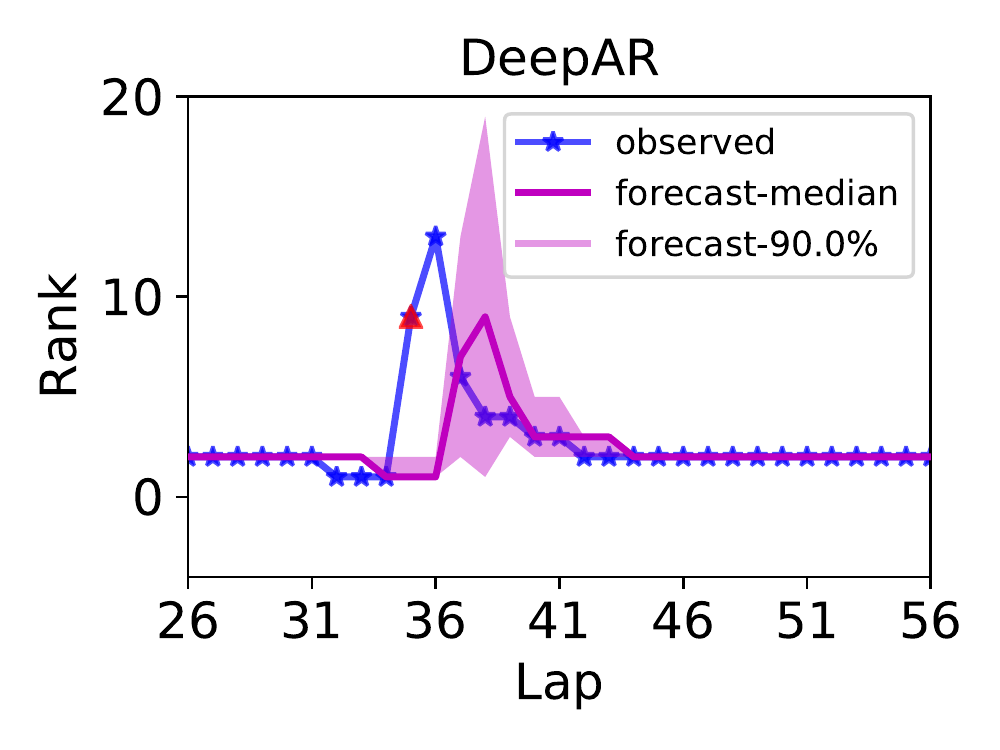}  
\caption{Two laps forecasting results around pit stop lap 34 for car\testcar \ in \testevent. (a)(b) Machine learning regression models. SVM learns a model very close to a two laps delay. RandomForest fails to predict the change around pit stop. (c) Statistical methods. ARIMA provides uncertainty predictions but lower performance, with difficulty to model the highly dynamics. (d) DeepAR, a state-of-the-art LSTM encoder-decoder model with uncertainty forecasting, also performs not well around pit stop.}
\label{fig:ranknet_vs_others}
\end{figure*}
\textbf{Rank position} is the order of the cars crossing SF/SFP, updated at each timeline during the race. 
Before the beginning of a race, the cars proceed forward row by row on the track. The start position of each car is assigned according to its performance in qualifying races. This stage is called the warm-up period. 
When the race begins, all cars start to accelerate, and the timing begins at the time when the first car crosses SF. Later on, each car's rank position of a lap is calculated by its elapsed time that finishing the lap.
In motorsports, a \textbf{pit stop} is a pause for refueling, new tires, repairs, mechanical adjustments, a driver change, a penalty, or any combination of them\citet{Pitstop_Understanding_nodate}.
Unexpected events happen in a race, including mechanical failures or a crash. Depending on the severity level of the event, sometimes it leads to a dangerous situation for other cars to continue the racing with high speed on the track. In these cases, a full course yellow flag rises to indicate the race entering a \textbf{caution laps} mode, in which all the cars slow down and follow a safety car and can not overtake until another green flag raised. 
\subsection{Rank position forecasting problem and challenges}
\label{sec:rank_problem}
The task of rank position forecasting is to predict the future rank position of a car given the race's observed history. 
Fig.\ref{fig:indycar_data}(a) shows the data collected by the sensors through the on-premises communication network.
Fig.\ref{fig:indycar_data}(b) shows a typical $Rank$ and $LapTime$ sequence. 
Both of them are stable most of the time, indicating the predictable aspects of the race that the driver's performance is stable. However, they both show abrupt changes when the \textbf{racing status}, including $LapStatus$ and $TrackStatus$, changes. 
Pit stops slow down the car and lead to a lose of rank position temporarily in the next few laps. Caution laps also slow down the car but do not affect the rank position much. 
Fig.\ref{fig:indycar_data}(b) demonstrates the characteristic of highly dynamic of this problem. \emph{The data sequence contains different phases, affected by the racing status.}
As for pit stop decisions, a team will have an initial plan for pit stops before the race, and the team coach will adjust it dynamically according to the status of the race. 'Random' events, such as mechanical failures and crashes, also make impacts the decision. 
A few laps of adjustment to the pit stop strategy may change the whole course of the race. However, when assuming the pit stop on each lap is a random variable,  only one realization of its distribution is observed in one race. 
Therefore, even the cause effect relationship between pit stop and rank position is known, forecasting of rank is still challenging due to the uncertainty in pit stop events.
These challenges, if not addressed accordingly, will hurt the forecasting performance of the model. Fig.\ref{fig:ranknet_vs_others} illustrates the limitations of typical existing methods. 
\begin{figure}[ht]
    \centering
    \includegraphics[width=0.5\linewidth]{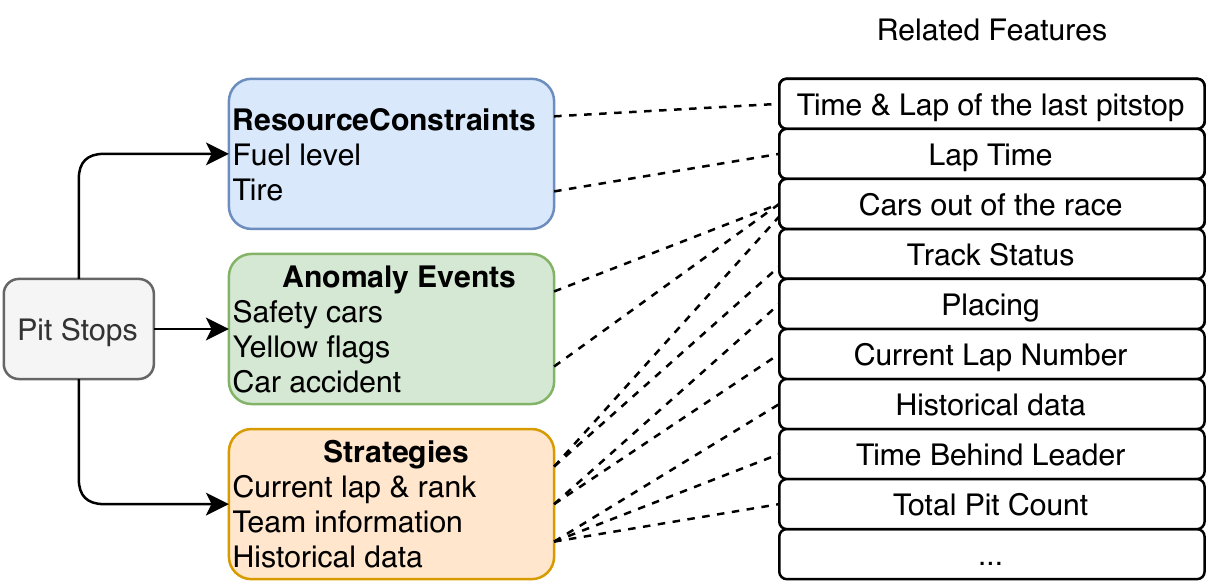}
    \caption{The main factors affecting Pit stop and their corresponding features.}
    \label{fig:pitstopfactors}
\end{figure}
\section{Models}
\label{sec:models}
\subsection{Pitstop factor analysis and related feature}
\label{sec:models-factor}
\begin{figure*}[ht]
\centering
\includegraphics[width=0.24\linewidth]{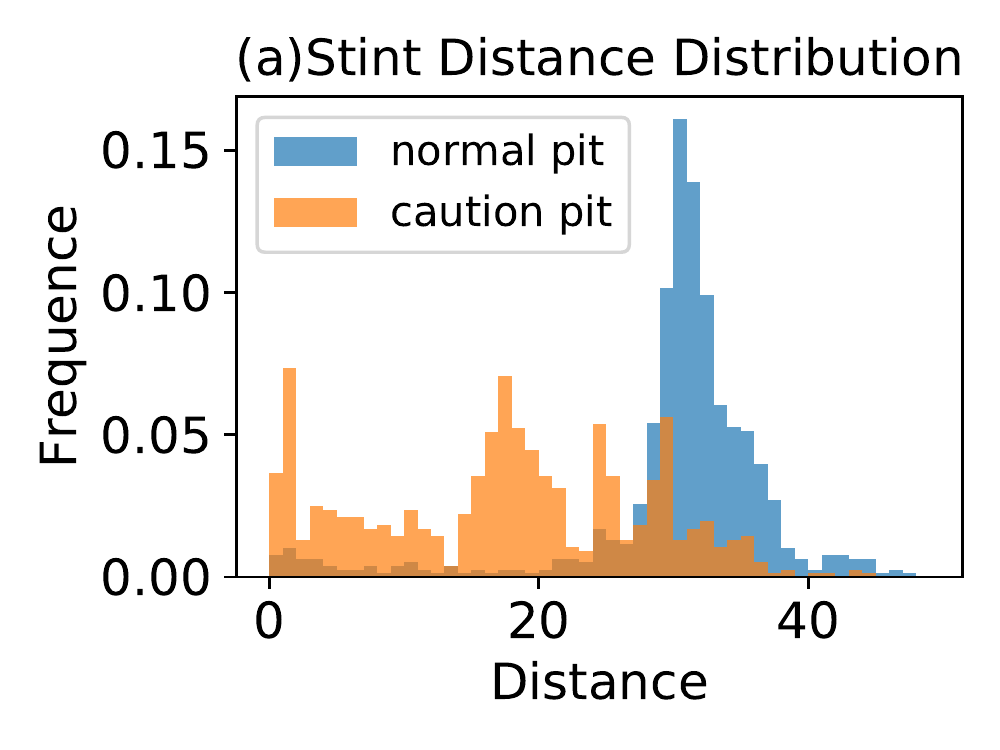} 
\includegraphics[width=0.24\linewidth]{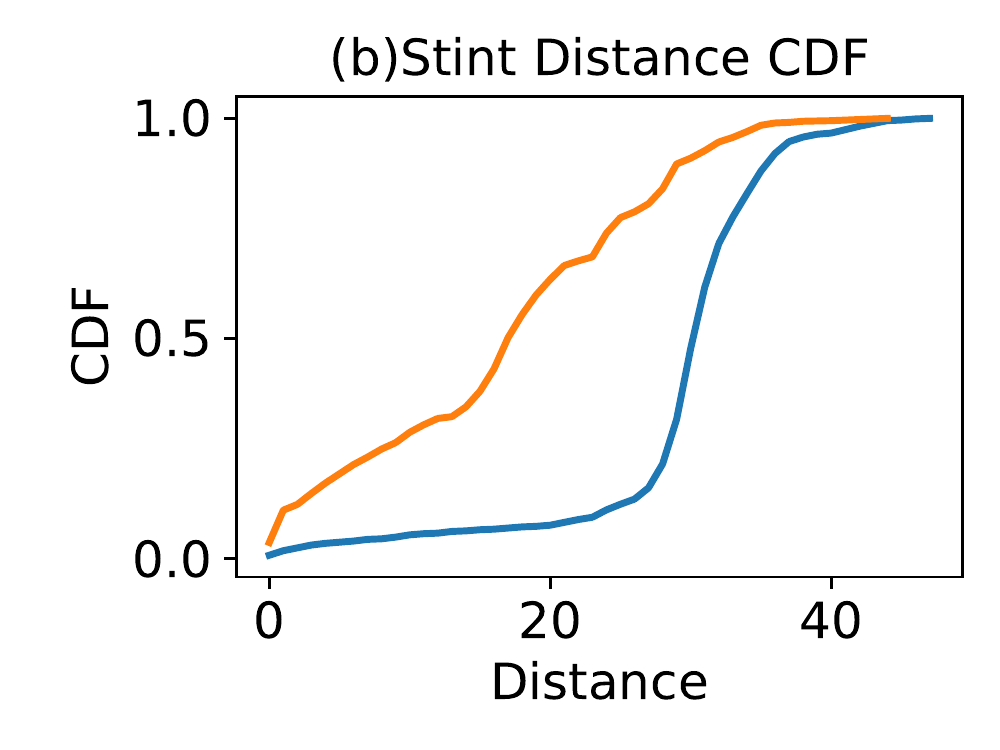} 
\includegraphics[width=0.24\linewidth]{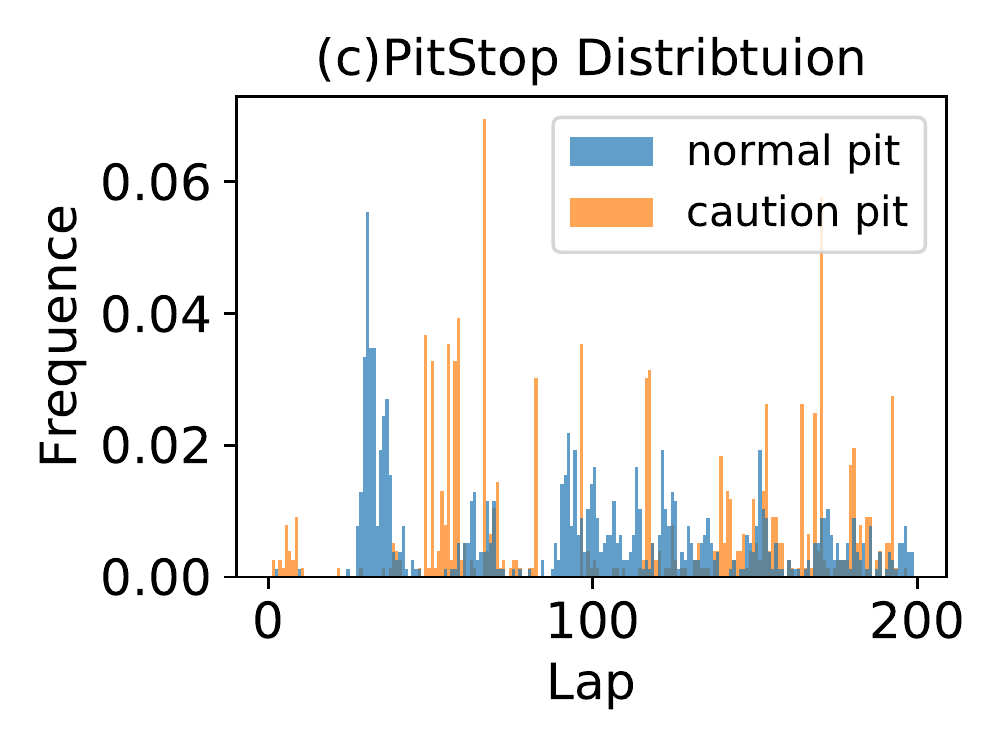} 
\includegraphics[width=0.24\linewidth]{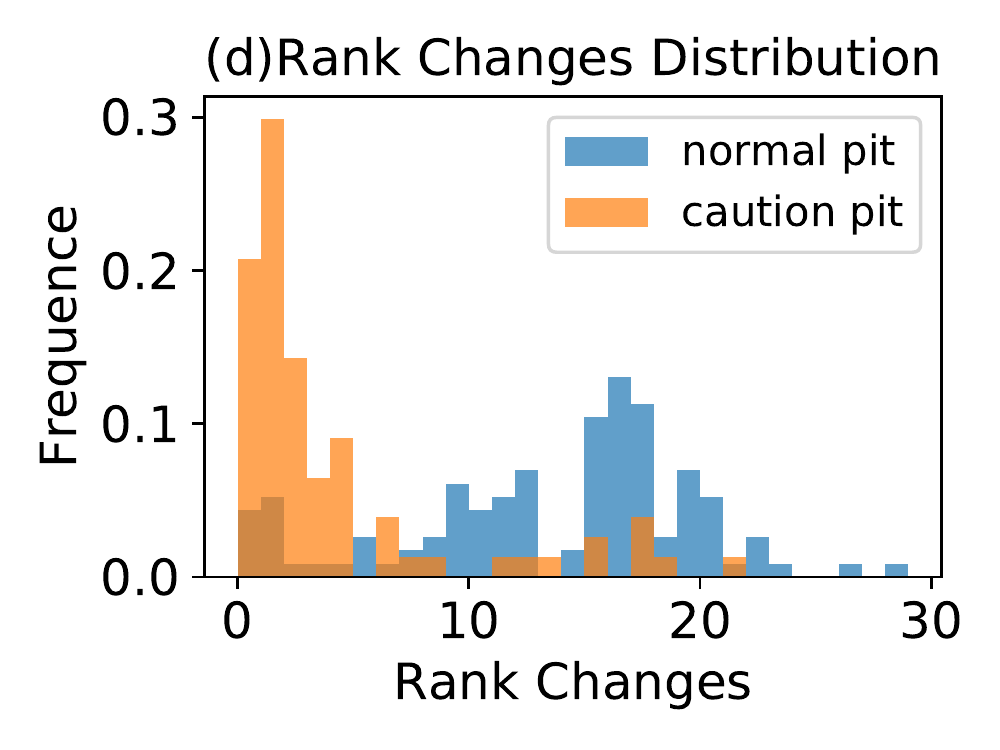} 
\caption{Statistics and analysis of pit stop. Stint refers to laps between two consecutive pit stops. Pit stops occurred on caution lap denoted as Caution Pit, otherwise Normal pit. (a)(b) Distribution of stint distance. Normal pits and caution pits are different. (c) Large uncertainty of where pit stops occur. (d) Caution  pits  has  much  less  impacts  on  rank  position compared with normal pits.}
\label{fig:pitmodel}
\end{figure*}
\begin{table*}[ht]
\centering
\small
\caption{Summary of features used in RankNet model}
\begin{tabular}{lllm{50ex}}
\hline
Variable   &Feature  &Domain    &Description  \\
 \hline
\multirow{5}{4em}{Race status $\mathbf{X}_i$ } & $TrackStatus(i,L)$  & T/F & Status of each lap for a car $i$, normal lap or caution lap.  \\
 &$LapStatus(i,L)$ &  T/F  & Whether lap $L$ is a pit stop lap or not for car $i$    \\
& $CautionLaps(i,L)$ & $\mathbb{N}$ & At Lap $L$, the count of caution laps since the last pit lap of car $i$. \\
& $PitAge(i,L)$ &$\mathbb{N}$ & At lap $L$, the count of laps after the previous pit stop of car $i$. \\
\hline
\multirow{3}{4em}{Rank $\mathbf{Z}_i$ } & $Rank(i,L)$  & $\mathbb{N}$ & There are $Rank(i,L)$ cars that completed lap $L$ before car $i$   \\
& $LapTime(i,L)$ &$\mathbb{R}+$  & Time used by car $i$ to complete lap $L$.   \\
& $TimeBehindLeader(i,L)$ &$\mathbb{R}+$  & Time behind the leader of car $i$ in lap $L$.   \\
\hline
\end{tabular}
\label{tab:feature}
\end{table*}
Previous studies\citet{heilmeier2018race, choo_real-time_2015,tulabandhula_tire_2014} did some preliminary analysis of the factors that affect pit stop. In this section, we study the causes of Pit stop based on the data of Indy500, which will help us select the main features and build a deep learning model. As in Fig.~\ref{fig:pitstopfactors}, we divide the causes of Pit Stop into three categories: \textbf{resource constraints}, \textbf{anomaly events}, and \textbf{human strategies}. 
\paragraph{Resource constraints}
The distance between the two pit stops is limited by the car's fuel tank volume and the car's tires. 
As in Fig.~\ref{fig:pitmodel}(a), no car run more than 50 laps before entering the pit stop.
\paragraph{Anomaly events}
Anomaly events are usually caused by mechanical failure or car accidents. When a serious accident occurs, $TrackStatus$ will change to Yellow Flag, which will change the pit stop strategy. 
In the Indy500 dataset, the number of the normal pit and caution pit are close, 777 and 763, respectively. These two types of pit stops show significant differences. In Fig. \ref{fig:pitmodel}(a), the normal pit is a bell curve, and caution pit scatters more evenly; In Fig. \ref{fig:pitmodel}(b), the CDF curve shows the lap distance of normal pit can be split into three sections, roughly from [0-23,24-40,40-]. The lower section of short distance pit may be caused mainly by unexpected reasons, such as mechanical failures, and keeps a low probability of less than 10\%. The upper part of the long-distance pit is mainly observed when a lot of caution laps happen, in which case the cars run at a reduced speed that greatly reduces tire wear and fuel burn for a distance traveled. From these observations, modeling pit stops on the raw pit data could be challenging, and modeling the normal pit data and removing the short distance section is more stable.
\paragraph{Human strategies}
In real competitions, participants also need to make decisions based on information such as the progress of the competition and the current ranking. In most cases, human strategies are very delicate and difficult to summarize with simple rules.
In order to better understand human strategy, we need to combine the ranking, team information, and historical data of past races to train the model.
\subsection{Modeling uncertainty in high dynamic sequences}
\begin{figure}[ht]
\centering
\subfloat[Process of forecasting. History data first feed into PitModel to get RaceStatus in the future, then feed into RankModel to get Rank forecasting. The output of the models are samples drawed from the learned distribution.The features contained in the vectors $\mathbf{X}_i$ and $\mathbf{Z}_i$ are shown in Table \ref{tab:feature}. Encoder length is set to $60$ in actual use.]{\includegraphics[width=0.45\linewidth]{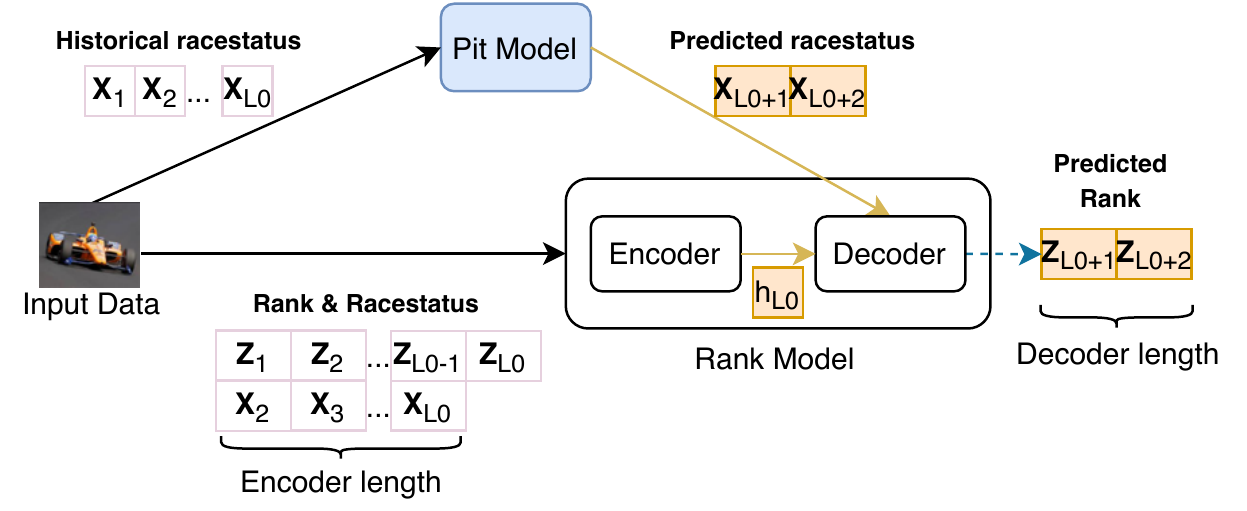}}
\hspace{3ex}
\subfloat[PitModel is a MLP predicting next pit stop lap given features of RaceStatus history. RankModel is stacked 2-layers LSTM encoder-decoder predicting rank for next prediction\_len laps, given features of historical Rank and RaceStatus, and future RaceStatus predicted by PitModel.]{\includegraphics[width=0.5\linewidth]{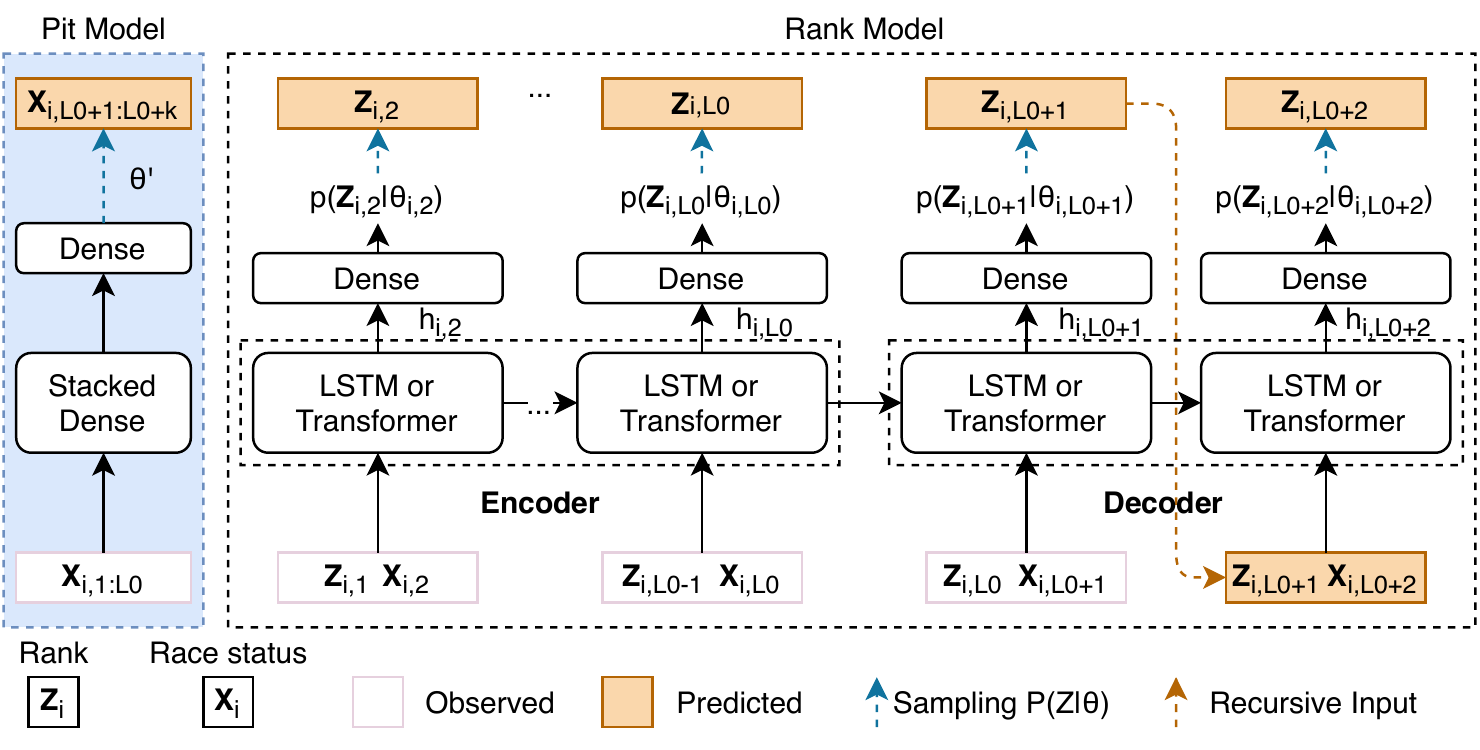}}
\caption{RankNet architecture.}
\label{fig:arch}
\end{figure}
\begin{algorithm}[ht]
\small
\caption{Training a minibatch of RankNet}
\label{alg:Training}
\SetKwInOut{Input}{input}
\SetKwInOut{Output}{output}
\Input{A minibatch ($batch\_size = B$) of time serises $\{z_{i,1:L_0+k} \}_{i=1,\dots B}$ and associated covariates $\{\mathbf{x}_{i,1:L_0+k} \}_{i=1,\dots B}$.}
\For{$i = 1 \dots B$ and $L=L_0+1 \dots L_0+k $ }
{
Calculate the current state $\mathbf{h}_{i,L} = h(\mathbf{h}_{i,L-1},z_{i,L-1},\mathbf{x}_{i,L}  )$ through the neural network.\\
Calculate the parameter $\theta_{i,L} = \theta (\mathbf{h}_{i,L})$ of the likelihood $p(z|\theta)$.
}
The loss function is obtained by log-likelihood:
\begin{equation}
\mathcal{L} = \sum_{i=1}^B \sum_{L=L_0+1}^{L_0+k} \log{p(z_{i,L}|\theta(\mathbf{h}_{i,L}))}
\end{equation}
\\
Apply the ADAM\citet{kingma2014adam} optimizer to update the weights of the neural network by maximizing the log likelihood $\mathcal{L}$.
\end{algorithm}
\begin{algorithm}[ht]
\small
\caption{Forecasting of RankNet}
\label{alg:Prediction}
\SetKwInOut{Input}{input}
\SetKwInOut{Output}{output}
\Input{$\{\mathbf{x}_{i,1:L0}\}, \{z_{i,1:L0}\}$ , model trained with prediction\_length $k$, forecasting start position $L_0$, end position $L_P$.}
Forecasting future pit stops by using the PitModel shown in Fig. \ref{fig:arch} (b) and calculate $\mathbf{x}_{L_0+1 : L_P}$; set future TrackStatus to zero.\\
\While{$L_0 < L_P$}{
Input the historical data at lap $L \le L_0$ into the RankModel to obtain the initial state $\mathbf{h}_{i,L_0}$.\\
\For{$L =L_0,...L_0+k-1$ }
{
Input $\{ z_{i,L}, \mathbf{x}_{i,L+1}, \mathbf{h}_{i,L} \}$ into the RankModel to get $\theta_{i,L+1}$\\
Random sampling  $\tilde z_{i,L+1}\sim p(\cdot|\theta_{i,L+1})$. \\
Update $z_{i,L+1}$ with $\tilde z_{i,L+1}$
}
$L_0 += k$\\
}
Return ~$\tilde z_{i,L_P}$
\end{algorithm}
We treat the rank position forecasting as a sequence-to-sequence modeling problem.
We use $z_{i,L}$ to denote the value of sequence $i$ at lap $L$, $\mathbf{x}_{i,L}$ to represent the covariate that are assumed to be known at any given lap. 
An encoder-decoder architecture is employed to map a input sequence $[z_{i,1},z_{i,2}, \dots z_{i,L_0}]$ to the output sequence $[z_{i,L_0+1} \dots z_{i,L_0+k}]$. Here $L_0$ represents the length of the input sequence, and $k$ represents the prediction length. Note that lap number $L$ is relative, i.e. $L=1$ corresponds the beginning of the input, not necessarily the first lap of the actual race.
To modeling the uncertainty, we follow the idea proposed in \citet{salinas_deepar_2019} to deliver probabilistic forecasting.
Instead of predicting value of target variable in the output sequence directly, a neural network predicts all parameters $\theta$ of a predefined probability distribution $p(z|\theta)$ by its output $h$. 
For example, to model a Gaussian distribution for real-value data, the parameter $\theta=(\mu,\sigma)$ can be calculated as: $\mu(h_{i,L})=W_\mu ^Th_{i,L} + b_\mu$, $\sigma(h_{i,L})=log(1+exp(W_\sigma^T h_{i,L} + b_\sigma))$. The final output $z_{i,L}$ is sampled from the distribution.
Our goal is to model the conditional distribution
$$P(z_{i, L_0+1: L_0+k} | z_{i,1:L_0}, \mathbf{x}_{i,1:L_0+k})$$
We assume that our model distribution $Q_{\Theta} (\mathbf{z}_{i, L_0+1: L_0+k} | z_{i,1:L_0}, \mathbf{x}_{i,1:L_0+k}) $ consists of a product of likelihood factors
\begin{equation}
\begin{aligned}
& Q_{\Theta} (\mathbf{z}_{i, L_0+1: L_0+k} | \mathbf{z}_{i,1:L_0}, \mathbf{x}_{i,1:L_0+k})\\  = &  \prod_{L=L_0+1}^{L_0+k}Q_{\Theta}(z_{i, L} | z_{i,1:L-1}, \mathbf{x}_{i,1:L_0+k}) \\
 = & \prod_{L=L_0+1}^{L_0+k}p(z_{i, L} | \theta (\mathbf{h}_{i,L}, \Theta) )
\end{aligned}
\end{equation}
parametrized by the output $h_{i,L}$ of an autoregressive recurrent
network
$$\mathbf{h}_{i,L} = h (\mathbf{h}_{i,L-1} ,z_{i,L-1}, \mathbf{x}_{i,L}, \Theta )$$
where $h$ is a function that is implemented by a multi-layer recurrent neural network with LSTM cells parametrized by $\Theta$.
\begin{table*}
\centering
\caption{Summary of the data sets.}
\small
\label{tbl:dataset}
\begin{tabular}{ccccccccc} 
\hline
Event & Year& Track Length &Track Shape & Total Laps & Avg Speed &\# Participants   & \# Records  &Usage \\
\hline
Indy500& 2013-2017 & 2.5& Oval &200 & 175mph &33    & 6600&Training  \\ 
Indy500& 2018 & 2.5& Oval &200 & 175mph &33    & 6600 &Validation  \\ 
Indy500& 2019 & 2.5& Oval &200 & 175mph &33    & 6600 &Test  \\ 
\hline
Iowa&  2013, 2015-2018  & 0.894& Oval&250&135mph& 21-24  & 6000 &Training\\ 
Iowa&  2019  & 0.894& Oval&300&135mph& 22  & 7200 & Test\\ 
\hline
Pocono&2013, 2015-2017  & 2.5 & Triangle &160 & 135mph & 22-24   &3840& Training\\ 
Pocono&2018  & 2.5& Triangle  &200 & 135mph & 22   &4800& Test \\ 
\hline
Texas & 2013-2017 & 1.455& Oval &228 & 153mph & 22-24   & 5472 & Training\\
Texas & 2018-2019 & 1.455 & Oval&248 & 153mph & 22   & 5704& Test\\
\hline
\end{tabular}
\end{table*}
The encoder-decoder architecture provides an advantage by supporting to incorporate covariates known in the forecasting period. For example, in sales demand forecasting, holidays are known to be important factors in achieving good predictions. These variables can be expressed as covariates inputs into both the encoder network and the decoder network. As we know, caution laps and pit stops are important factors to the rank position. Therefore, they can be considered as covariate inputs. But, different from the holidays, these variables in the future are unknown at the time of forecasting, leading to the need to decompose the cause effects in building the model.
\subsection{Modeling extreme events and cause effects decomposition}
Changes of race status, including pit stops and caution laps, cause the phase changes of the rank position sequence.
As a direct solution to address this cause effect, we can model the race status and rank position together and joint train the model in the encoder-decoder network.
In this case, target variable $z_{i,t}$ is a multivariate vector $[Rank, LapStatus, TrackStatus]$. 
However, this method fails in practice due to data sparsity. The changes of race status are rare events, and
targets of rare events require different complexity of models. 
For example, on average a car goes to pit stop for six times in a race. Therefore, LapStatus, a binary vector with length equals to 200, contains only six ones, 3\% effect data. 
TrackStatus, indicating the crash events, is even harder to predict.
We propose to decompose the cause effect of race status and rank position in the model. 
RankNet, as shown in Fig. \ref{fig:arch}(a), is composed with two sub-models. First, a PitModel forecasts the future RaceStatus, in which LapStatus is predicted and  
TrackStatus is set to zeros assuming no caution laps in the future.
Then the RankModel forecasts the future Rank sequence. 
RaceStatus is the most important feature in covariates $X_{t}$. $TrackStatus$ indicates whether the current lap is a caution lap, in which the car follows a safety car in controlled speed. $LapStatus$ indicates whether the current lap is a pit stop lap, in which the car cross SF/SFP in the pit lane.
Some other static features can also be added into the input. For example, CarId represents the skill level of the driver and performance of the car.
Transformations are applied on these basic features to generate new features. Embedding for categorical CarId is utilized. Accumulation sum transforms the binary status features into 'age' features, generating features such as CautionLaps 
and PitAge. Table \ref{tab:feature} summarizes the definition of these features. 
 For efficiency, instead of sequences input and output, PitModel in Fig.\ref{fig:arch}(b) use CautionLaps and PitAge as input, and output a scalar of the lap number of the next pit stop.
A rank position forecasting network is trained with a fixed prediction length. 
In order to deliver a variable length prediction, e.g., in predicting the rank positions between two pit stops, we apply a fixed length forecasting regressively by using previous output as input for the next prediction. 
For the probabilistic output, we take 100 samples for each forecasting. And the final rank positions of the cars are calculated by sorting the sampled outputs. 
The training and prediction process of the model is shown in Algorithm \ref{alg:Training} and Algorithm \ref{alg:Prediction}.
\section{Experiments}
\subsection{Dataset}
\begin{figure}[ht]
\centering
 \includegraphics[width=0.5\linewidth]{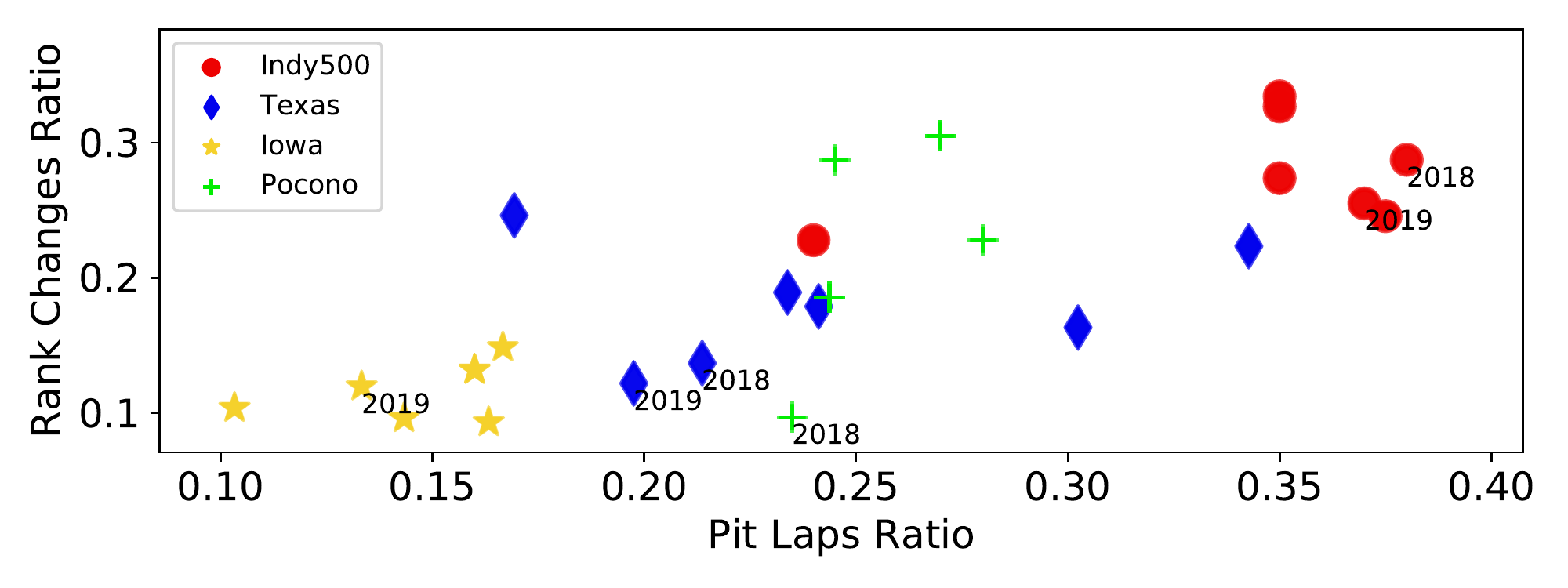}
\caption{Data distribution of Indycar Dataset. PitLapsRatio is the pit stop laps \# divided the total laps \#. RankChangesRatio refers to the ratio of laps with rank position changes between consecutive laps.}
\label{fig:dataset}
\end{figure}
\begin{figure*}[ht]
\centering
\includegraphics[width=1\linewidth]{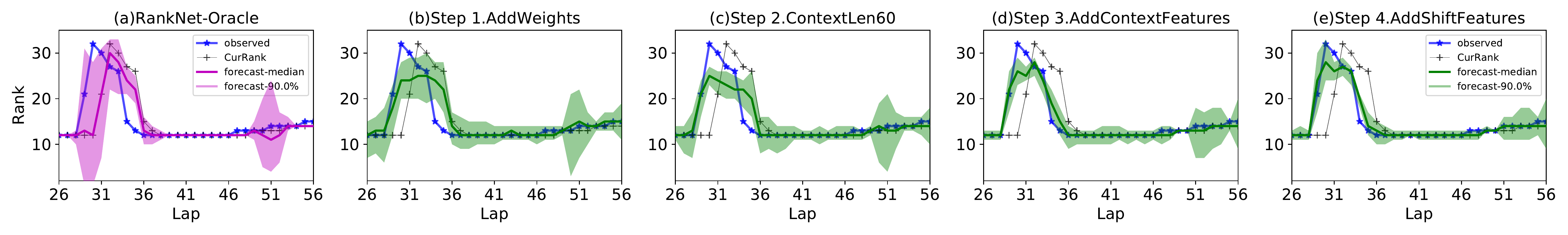}
\caption{Illustration of RankNet model optimization on two laps forecasting for Car13 Indy500-2018.(a)Basic RankNet model trained with Oracle race status features and context\_length=40. (b)Adding larger weights to the loss for instances with rank changes, set optimal weight to 9. (c)Tuning on parameter context\_length, set optimal length to 60. (d)Adding context features, including LeaderPitCount: \# of leading cars(based on the rank position at lap A-2) that go to pit stop at lap A; TotalPitCount:\# of cars that go to pit stop at lap A. (e)Adding shift features, including Shift\_RaceStatus: lapstatus and trackstatus of the future at lap A+2;
Shift\_TotalPitCount:\# of cars that go to pit stop at lap A+2.}
\label{fig:modelopt}
\end{figure*}
We evaluate our model on the car racing data of IndyCar series\citet{IndyCar_Dataset_nodate}. 
Due to the data scarcity in car racing,
we have to incorporate more data to learn a stable model. 
Using the historical data that is too long time ago can be ineffective because many factors change along the time, including the drivers' skills, configurations of the cars, and even the rules of the race. The same year data of other races are 'similar' in the status of the drivers, cars, and rules, but different shapes and lengths of the track lead to different racing dynamics.
In this paper, we select races of superspeedway after 2013 with at least 5 years data each, and after removing corrupted data, get a dataset of 25 races from four events, shown in Table.~\ref{tbl:dataset}.
Fig.~\ref{fig:dataset} shows the data distribution by two statistics for this dataset. Among all the events, Indy500 is the most dynamic one which has both the largest PitLapsRatio and RankChangesRatio, Iowa is the least. 
We train models separately for each event. Races of the first five years are used as the training dataset, the remains are used as testing data, which are labeled in Fig.~\ref{fig:dataset}. Since Pocono has only five years of data in total, its training set uses four of them. 
First, we start from Indy500 and use Indy500-2018 as validation set. Then we investigate the generalization capability of the model on data of the other events.
\subsection{Baselines and implementation}
\begin{table}
\centering
\caption{Features of the rank position forecasting models.}
\label{tab:model-features}
\small
\begin{tabular}{l|ccc} 
\hline
Model          & Representation & Uncertainty & PitModel          \\ 
          & Learning & Support  &  Support \\ 
\hline
Random Forest & N    & N & N   \\ 
SVM & N    & N & N   \\ 
XGBoost & N    & N & N   \\ 
ARIMA          & N & Y  & N  \\ 
DeepAR         & Y & Y & N     \\ 
RankNet-Joint  & Y & Y & Y(Joint Train)    \\ 
RankNet-MLP    & Y & Y & Y(Decomoposition) \\
RankNet-Oracle    & Y & Y & Y(Ground Truth) \\
\hline
\end{tabular}
\end{table}
As far as we know, there is no open source model that forecast rank position in car racing, and no related work on IndyCar series.
First, we have a naive baseline which assumes that the rank positions will not change in the future, denoted as CurRank. 
Secondly, We implemented machine learning regression models as baselines that follow the ideas in \citet{tulabandhula_interactions_2014} which forecast change of rank position between two consecutive pit stops, including RandomForest, SVM and XGBoost that do point wise forecast. 
DeepAR\citet{salinas_deepar_2019} is one of the state-of-the-art deep time series models, it uses the same network architecture as in Fig.~\ref{fig:arch}(c) without covariates of TrackStatus and LapStatus. 
Depending on the approach of dealing with pit stop modeling, RankNet has three implementations: 1. RankNet-Joint is the model that train target with pit stop jointly without decomposition.
2. RankNet-Oracle is the model with ground truth TrackStatus and LapStatus as covariates input. It represents the best performance that can be obtained from RankNet given the caution and pit stop information for a race. 
3. RankNet-MLP deploys a separate pit stop model, which is a multilayer perceptron(MLP) network with probability output, as in Fig.~\ref{fig:arch}(b).  
Table.~\ref{tab:model-features} summarizes the features of all the models.
We build our model RankNet with the Gluonts framework\citet{alexandrov_gluonts_2019}. Gluonts is a library for deep-learning-based time series modeling, enables fast prototype and evaluation. RankNet is based on the DeepAR implementation in Gluonts, shares the same features, including: sharing parameters between encoder and decoder, encoder implemented as a stacking of two LSTM layers. 
\subsection{Model optimization}
\begin{table}[ht]
\centering
\caption{Dataset statistics and model parameters}
\small
\begin{tabular}{l|l}
\hline
Parameter & Value\\
\hline
\# of time series & 227(Indy500), 619(All)\\
\# of training examples & 32K(Indy500), 117K(All)\\
Granularity & Lap \\
Domain & $\mathbb{R}+$ \\
Encoder length $C = L_0-1$& \textbf{60}\\
Decoder length $k$ & 2\\
Loss weight & [1-10]\\
Batch size $B$& \textbf{32},64,128,256,640,1600, 3200 \\
Optimizer & ADAM \\
Learning rate & 1e-3\\
LR Decay Factor & 0.5 \\
\# of lstm layers & 2\\
\# of lstm nodes & 40\\
Training time & 2h\\
\hline
\end{tabular}
\label{tab:model:para}
\end{table}
For machine learning baselines, the hyper-parameters are tuned by grid search.  For deep models, we tune the parameter of encoder length, loss weight and use the default value of other hyper-parameters in the GluonTs implementation, as in Table.~\ref{tab:model:para}. 
The model is trained by ADAM optimizer with early stopping mechanism that decay the learning rate when loss on validation set does not improve for 10 epochs until reaching a minimum value.
Fig.~\ref{fig:modelopt} shows the process of further model optimization, starting from a basic RankNet-MLP model, optimizations are added step by step and tuned on the validation dataset. 
\begin{table*}
\centering
\caption{Short-term rank position forecasting(prediction leghth=2) of Indy500-2019}
\label{tbl:rank_forecasting_shortterm}
\resizebox{\textwidth}{!}{
\begin{tabular}{l|cccc|cccc|cccc} 
\hline
Dataset & \multicolumn{4}{c|}{All Laps}       & \multicolumn{4}{c|}{Normal Laps}           & \multicolumn{4}{c}{PitStop Covered Laps}          \\ 
\hline
Model   & Top1Acc & MAE & 50-Risk     & 90-Risk     & Top1Acc    & MAE & 50-Risk     & 90-Risk     & Top1Acc    & MAE & 50-Risk     & 90-Risk      \\
\hline
CurRank & 0.73 & 1.16 & 0.080 & 0.080 & 0.94 & 0.13 & 0.009 & 0.009 & 0.55 & 2.09 & 0.144 & 0.144  \\ 
ARIMA   & 0.54 & 1.45 & 0.110 & 0.105 & 0.70 & 0.47 & 0.047 & 0.042 & 0.40 & 2.32 & 0.166 & 0.162  \\ 
RandomForest   & 0.62 & 1.31 & 0.091 & 0.091 & 0.78 & 0.39 & 0.027 & 0.027 & 0.47 & 2.14 & 0.147 & 0.147  \\ 
SVM     & 0.73 & 1.16 & 0.080 & 0.080 & 0.94 & 0.13 & 0.009 & 0.009 & 0.55 & 2.09 & 0.144 & 0.144  \\ 
XGBoost & 0.64 & 1.25 & 0.086 & 0.086 & 0.76 & 0.27 & 0.019 & 0.019 & 0.54 & 2.12 & 0.146 & 0.146  \\ 
DeepAR  & 0.73 & 1.22 & 0.086 & 0.085 & 0.93 & 0.21 & 0.018 & 0.017 & 0.55 & 2.12 & 0.147 & 0.145  \\ 
\hline
RankNet-Joint  & 0.64 & 1.74 & 0.153 & 0.144 & 0.78 & 0.82 & 0.096 & 0.089 & 0.52 & 2.56 & 0.203 & 0.194  \\ 
RankNet-MLP     & \textbf{0.79} & \textbf{0.94} & \textbf{0.067} & \textbf{0.046} & \textbf{0.94} & \textbf{0.21} & \textbf{0.022} & \textbf{0.018} & \textbf{0.65} & \textbf{1.59} & \textbf{0.107} & \textbf{0.072}  \\
RankNet-Oracle & 0.90 & 0.57 & 0.045 & 0.040 & 0.94 & 0.20 & 0.021 & 0.018 & 0.87 & 0.90 & 0.067 & 0.061  \\ 
\hline
\end{tabular}
}
\end{table*}
\subsection{Evaluation}
RankNet is a single model that able to forecast both short-term rank position(TaskA) and long term change of rank position between pitstops(TaskB). 
First, we use Mean Absolute Error(MAE) to evaluate average forecasting accuracy of all the sequences since they have the same units.  
Secondly, TaskA evaluates the accuracy of correct predictions of the leader, denoted as Top1Acc. TaskB evalutes the accuracy of correct predictions of the sign of the change which indicating whether a car achieves a better rank position or not, denoted as SignAcc.
Thirdly, a quantile based error metric $\rho$-risk\citet{seeger_bayesian_2016} is used to evaluate the performance of a probabilistic forecasting. 
When a set of samples output by a model, the quantile $\rho$ value of the samples is obtained, denoted as $\hat{Z_\rho}$, then $\rho$-risk is defined as $2(\hat{Z_\rho} - Z)((Z<\hat{Z_\rho}) - \rho)$, normalized by $\sum{Z_i}$. It quantifies the accuracy of a quantile $\rho$ of the forecasting distribution. 
\begin{figure*}
\centering
 \includegraphics[width=0.24\linewidth]{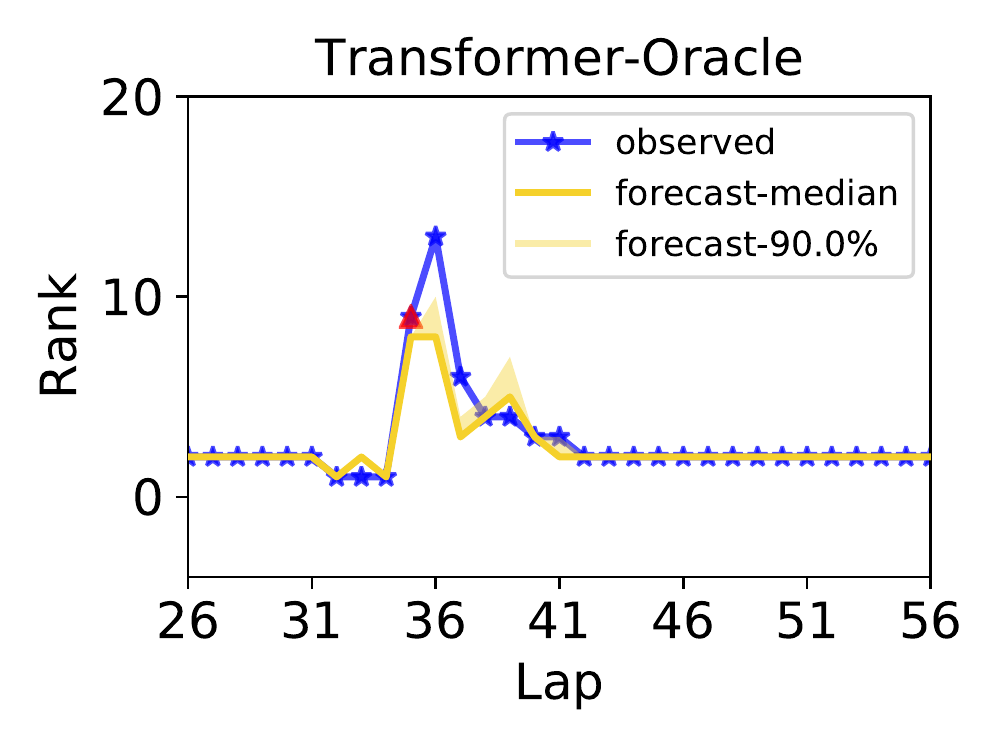} 
 \includegraphics[width=0.24\linewidth]{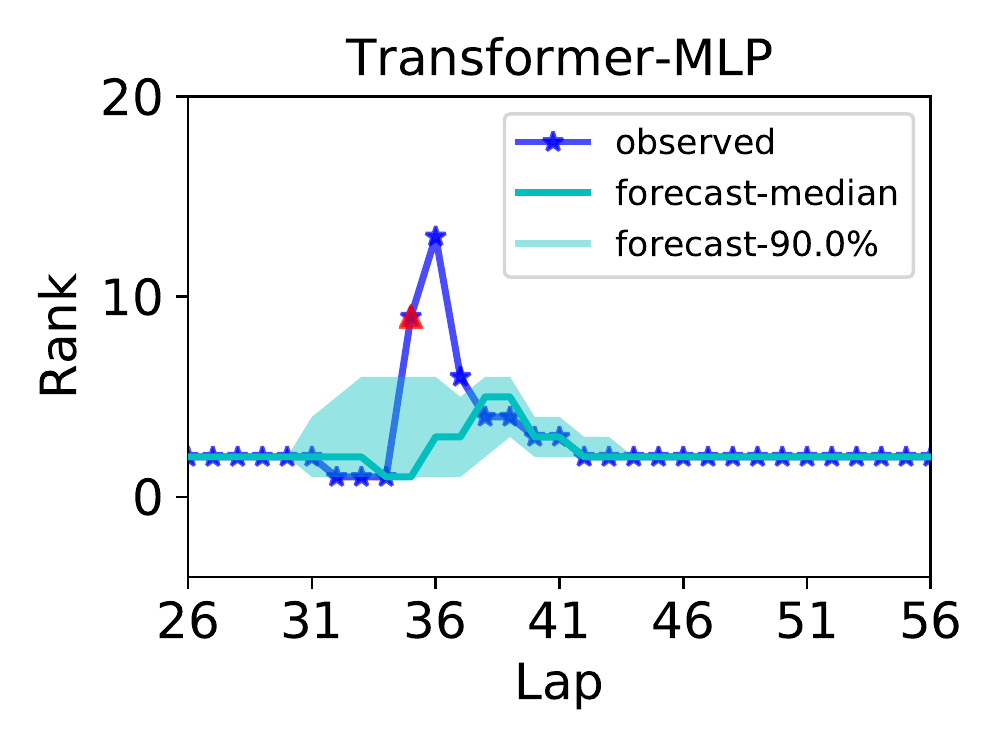} 
 \includegraphics[width=0.24\linewidth]{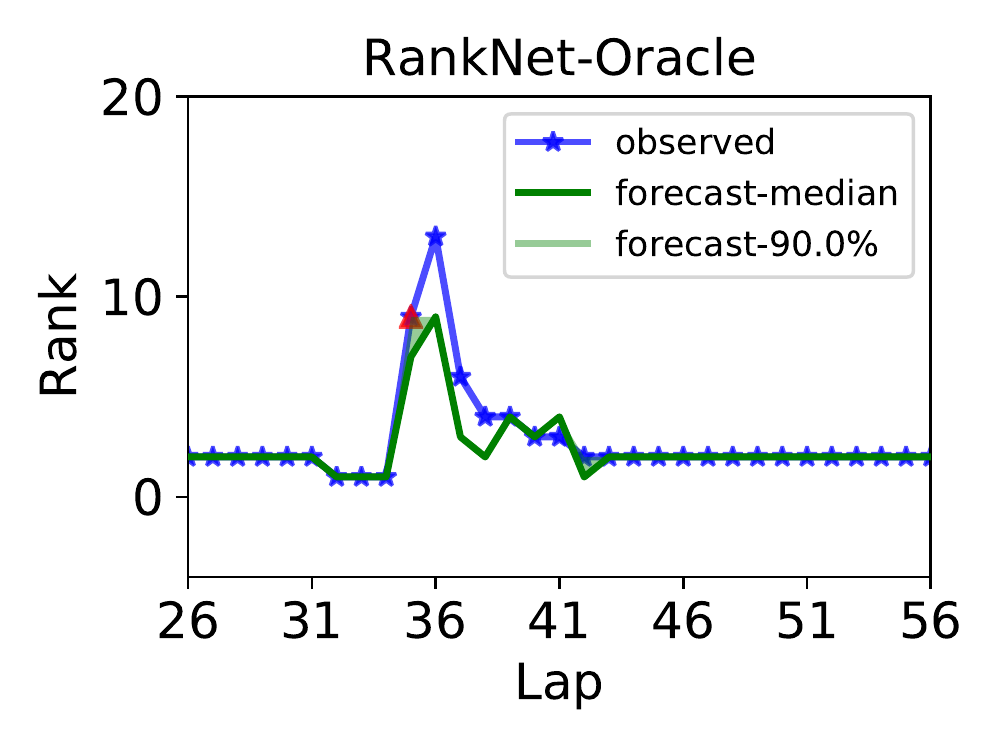} 
 \includegraphics[width=0.24\linewidth]{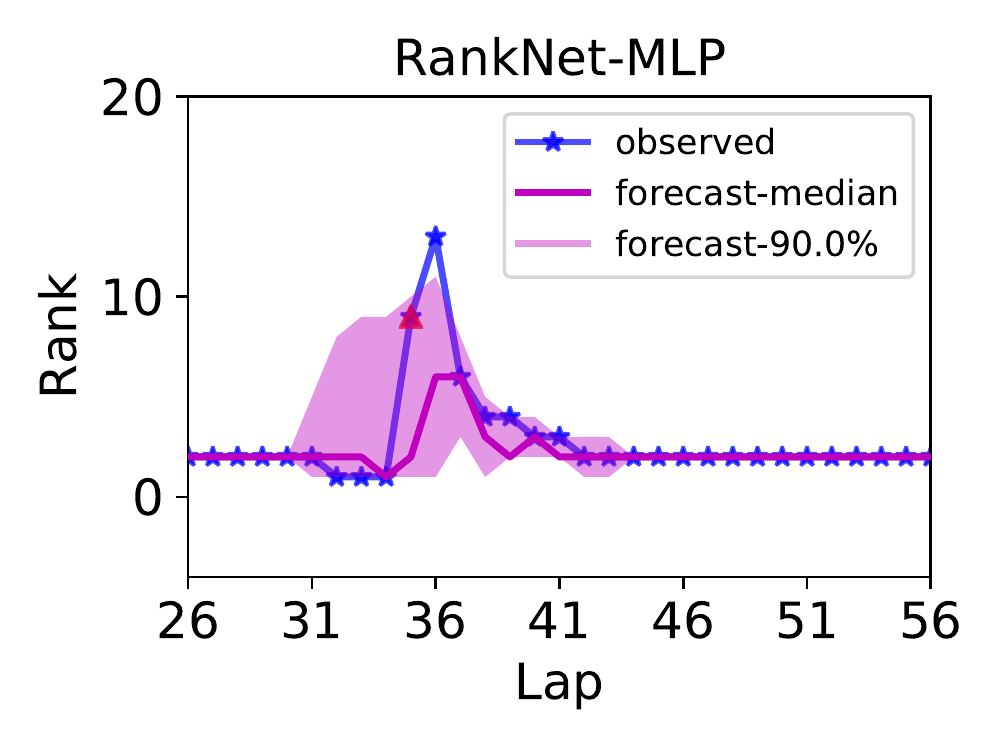} 
\caption{RankNet forecasting results of two laps in the future for car\testcar \ in \testevent.}
\label{fig:ranknet_vs_others_part2}
\end{figure*}
\subsection{Short-term rank position forecasting}
Table \ref{tbl:rank_forecasting_shortterm} shows the evaluation results of four models in a two laps rank position forecasting task. 
CurRank demonstrates good performance. 73\% leader prediction correct and 1.16 mean absolute error on Indy500-2019 indicates that the rank position does not change much within two laps. 
DeepAR is a powerful model but fails to exceed CurRank, which reflects the difficulty of this task that the pattern of the rank position variations are not easy to learn from the history. 
Other machine learning models, and RankNet-Joint all failed to get better accuracy than CurRank.
By contrast, RankNet-Oracle achieves significant better performance than CurRank, with 23\% better in Top1Acc and 51\% better in MAE. 
RankNet-MLP, our proposed model, is not as good as RankNet-Oracle, but still able to exceed CurRank by 7\% in Top1Acc and 19\% in MAE. It also achieves more than 20\% improvement of accuracy on 90-risk when probabilistic forecasting gets considered. 
Evaluation results on PitStop Covered Laps, where pit stop occurs at least once in one lap distance, show the advantages of RankNet-MLP and Oracle come from their capability of better forecasting in these extreme events areas.
A visual comparison of RankNet over the baselines are demonstrated in Fig.\ref{fig:ranknet_vs_others_part2} and Fig.\ref{fig:ranknet_vs_others}.
\subsection{Impact of prediction length}
\begin{figure}
\centering
\includegraphics[width=.5\linewidth]{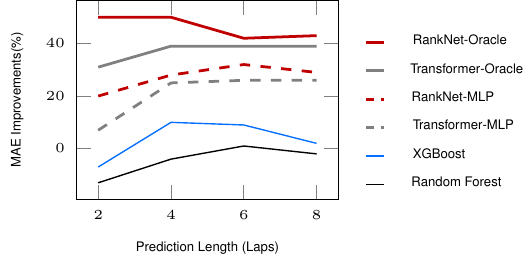}
\caption{Impact of prediction length over performance on Indy500-2019. Comparison  LSTM with Transformer for RankNet implementation on Indy500-2019.}
\label{fig:rank_plen}
\label{fig:transformer_plen}
\end{figure}
Fig.~\ref{fig:rank_plen} shows the impact of prediction length on relative forecasting performance. With the increasing of prediction length, the accuracy of all the models drops, and among all the models, RankNet-MLP and RankNet-Oracle keep a consistent MAE improvement over CurRank above 20\% and 40\% respectively. Models inferior to CurRank are not presented.
\begin{table}[ht]
\centering
\small
\caption{Rank position changes forecasting between pit stops}
\label{tbl:rank_change}
\begin{tabular}{l|cccc} 
\hline
Model          &  SignAcc &  MAE &  50-Risk &  90-Risk  \\ 
\hline
CurRank        & 0.15 & 4.33 & 0.280    & 0.262 \\ 
RandomForest   & 0.51 & 4.31 & 0.277    & 0.276 \\ 
SVM            & 0.51 & 4.22 & 0.270    & 0.249 \\ 
XGBoost        & 0.45 & 4.86 & 0.313    & 0.304 \\ 
DeepAR         & 0.37 & 4.08 & 0.265    & 0.268 \\ 
\hline
RankNet-Joint  & 0.60 & 5.83 & 0.388    & 0.486 \\ 
RankNet-MLP    &  \textbf{0.65} &  \textbf{3.79} &  \textbf{0.245}    &  \textbf{0.169} \\
RankNet-Oracle & 0.67 & 3.41 & 0.229    & 0.203 \\ 
\hline
\end{tabular}
\end{table}
\begin{table*}
\centering
\caption{Two laps forecasting task on other races. MAE improvements is compared over CurRank on PitStop covered laps.}
\label{tbl:eval_newraces}
\small
\begin{tabular}{l|cccc|cccc} 
\hline
   & \multicolumn{4}{c|}{MAE Improvement(Train by Indy500)}    & \multicolumn{4}{c}{MAE Improvement(Train by same event)} \\ 
\hline
Dataset      & \multicolumn{1}{c}{RankNet} & \multicolumn{1}{c}{Random} & \multicolumn{1}{c}{RankNet} & \multicolumn{1}{c|}{Transformer} & \multicolumn{1}{c}{RankNet}       & \multicolumn{1}{c}{Random}& \multicolumn{1}{c}{RankNet}      & \multicolumn{1}{c}{Transformer}        \\ 
             & \multicolumn{1}{c}{-MLP} & \multicolumn{1}{c}{Forest} & \multicolumn{1}{c}{-Joint} & \multicolumn{1}{c|}{-MLP} & \multicolumn{1}{c}{-MLP}       & \multicolumn{1}{c}{Forest}& \multicolumn{1}{c}{-Joint}      & \multicolumn{1}{c}{-MLP}        \\ 
\hline
Indy500-2019 & 0.24& -0.02& -0.08    & 0.12    & 0.24      & -0.02& -0.08& 0.12\\ 
Texas-2018   & 0.11& -2.13& -0.22    & 0.02    & 0.15      & -0.10& -0.11& 0.07   \\ 
Texas-2019   & 0.01& -1.63& -0.29    & -0.15   & 0.10      & -0.13& -0.15& -0.02 \\ 
Pocono-2018  & 0.09& -2.25& -0.02    & -0.17   & 0.06    & -1.51& -0.09& 0.02\\ 
Iowa-2019    & 0.09& -1.03& -0.09    & 0.03    & 0.09      & 0.09 & -0.07& 0.05\\
\hline
\end{tabular}
\end{table*}
\subsection{Stint rank position forecasting}
Table \ref{tbl:rank_change} shows the results on TaskB, forecasting the rank position changes between consecutive pit stops. 
CurRank can not predict changes, thus gets the worst performance. 
Among the three machine learning models, SVM shows the best performance.
RankNet-Oracle demonstrates its advantages over all the machine learning models, indicating that once the pit stop information is known, long term forecasting through RankNet is more effective. The performance of RankNet-MLP obtains significantly better accuracy and improves between 9\% to 30\% on the four metrics over SVM.
Moreover, it forecasts future pit stops and thus different possibilities of race status, which are not supported by the other baselines. RankNet is promising to be used as a tool to investigate and optimize the pit stop strategy. 
\subsection{Generalization to new races}
In the left column of Table.\ref{tbl:eval_newraces}, the models are trained by Indy500 training set, then tested on other race data. In the right column, the models are trained by the training set from the same event. 
RandomForest, as a representative of machine learning method, has its performance drops badly in the left column, indicating its incapability of adapting to the new data. 
On the contrary, RankNet-MLP obtains decent performance even when testing on unseen races.
It shows the advantages of RankNet model on generalizing to race data from different data distribution.
Pocono-2018 is a special case where RankNet-MLP trained by Pocono is worse than model trained by Indy500. As in Fig.~\ref{fig:dataset}, Pocono-2018 has small RankChangesRatio where CurRank delivers good performance; moreover, Pocono-2018 has the largest RankChangesRatio distance to other races from the same event, which makes it harder in forecasting with the trained model by the other races in Pocono. 
\subsection{RankNet with Transformer}
RankNet utilize the encoder-decoder architecture, where stacked LSTM or Transformer\citet{vaswani_attention_2017} can be used for the implementation of encoder and decoder network. In this experiment, we replace LSTM-based RNN with the Transformer implementation from GluonTs library, which has multi-head attention(8 heads) and dimension of the transformer network is 32. 
As in Fig.~\ref{fig:transformer_plen} and Table.~\ref{tbl:eval_newraces}, LSTM based RankNet demonstrates consistently a slightly better performance over Transformer based implementation.  We speculate that this is due to the small data size in our problem which limits Transformer to obtain better performance.
\subsection{RankNet Model Training Efficiency Evaluation}
When considering the deployment of RankNet in the field, continuous learning by keeping updating the model with newest racing data would be appropriate. In this section, we study the efficiency aspects in modeling training, answer the following questions: 
1. What challenges exist to accelerate the training process?
2. Which device is preferred to this need, including CPU, GPU, and Vector Engine (VE)? 
We re-implement RankNet with Tensorflow which is supported by all the devices, and conducted performance evaluation of the model training with hardware configuration in Table \ref{table:perf_hardware}.
\begin{table}
\centering
\caption{Experiments hardware specification.}
\small
\label{table:perf_hardware}
\begin{tabular}{l | c }
\hline
Platform & Hardware \\
\hline
CPU & Intel Xeon E5-2670 v3@2.30GHz,128G RAM \\
\hline
CPU+GPU &  Intel Xeon CPU E5-2630 v4,128G RAM\\
        &  GPU (V100-SXM2-16GB) \\
\hline
CPU+VE & Intel Xeon Gold 6126@2.60GHz,192G RAM\\
       & VE (SX-Aurora Vector Engine) \\
\hline
\end{tabular}
\end{table}
\begin{figure}[ht]
\centering
\includegraphics[width=.5\linewidth]{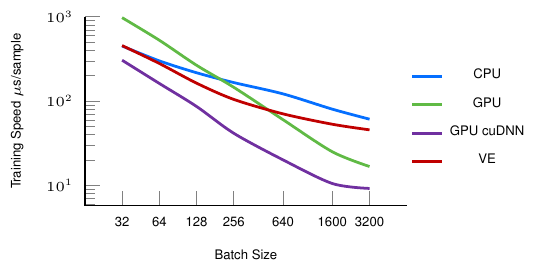}
\caption{Impact of batch\_size over training speed \begin{math}\mu\end{math}s/sample.}
\label{fig:impact_BatchSize}
\end{figure}
RankNet is a relative simple model with less than 30K parameters. 
As a RNN, data dependency between consecutive time step prevents parallelism cross time steps. Input to the model at each time step has the shape of batch\_size x feature\_dimension(a number less than 40, depends on how many transformations and regressive lags are employed.). It is hard to fully utilize the available computation resources with small input and model. We address this issue by increasing the batch\_size. Figure \ref{fig:impact_BatchSize} shows that the performance of training speed improve on all the configurations. With increasing the batch\_size, VE starts to perform better than CPU, while GPU's performance improvement is the largest. 
We also observe a much higher CPU utilization with the sustained peak close to 60\%. 
On the other side, batch\_size also has impacts on model convergence and accuracy, e.g., model trained with large batch\_size=3200 (under a larger learning rate) obtains the same level of validation loss and accuracy by using about $4\times$ epochs than the one with small batch\_size=32. Considering the large gains of training speed, as shown in Fig.~\ref{fig:impact_BatchSize}, where large batch\_size=3200 is more than $10\times$ faster, selecting the large batch\_size training is proved to be effective.
\begin{figure}[ht]
  \begin{center}
    \includegraphics[width=0.5\linewidth]{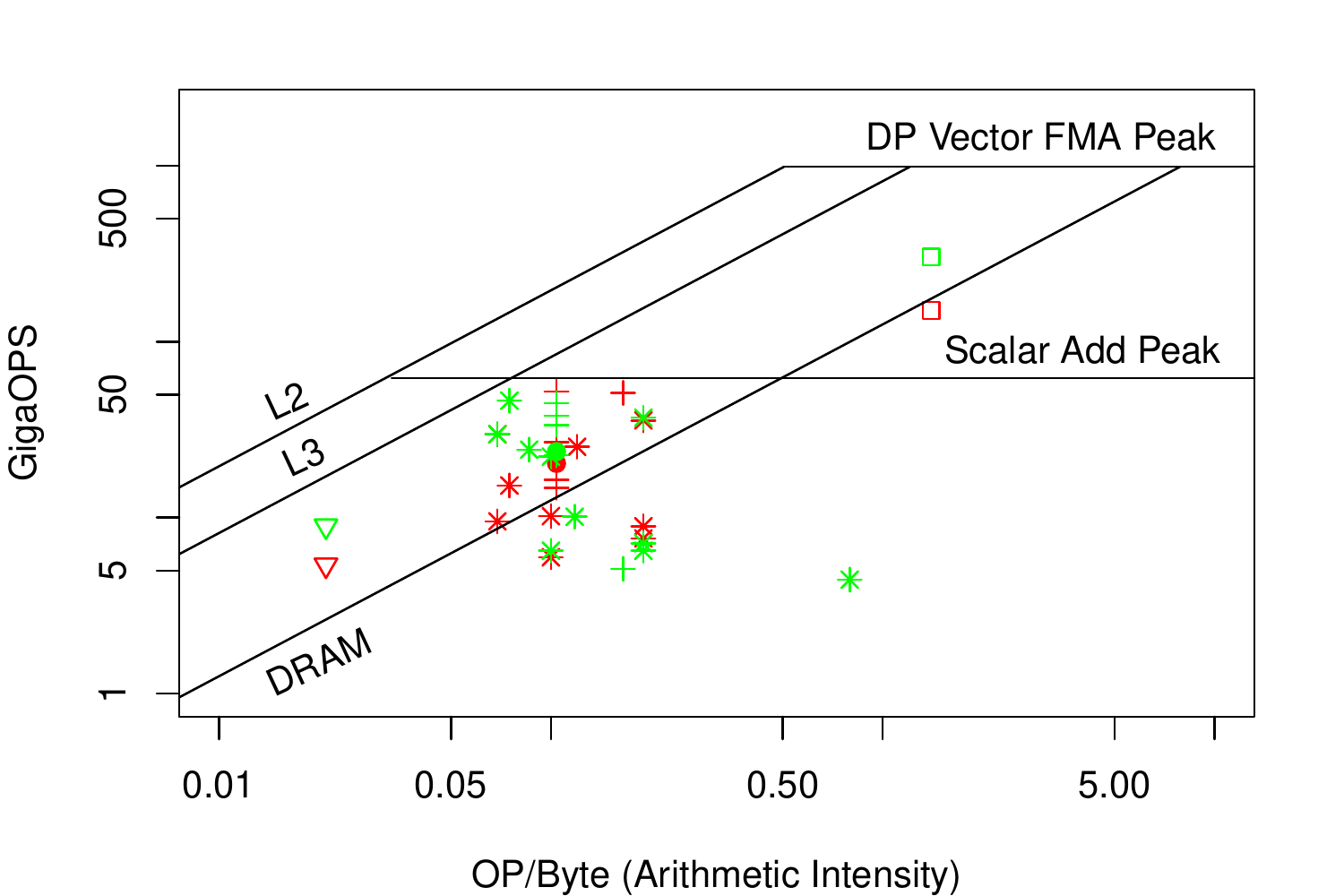}
    \includegraphics[width=0.5\linewidth]{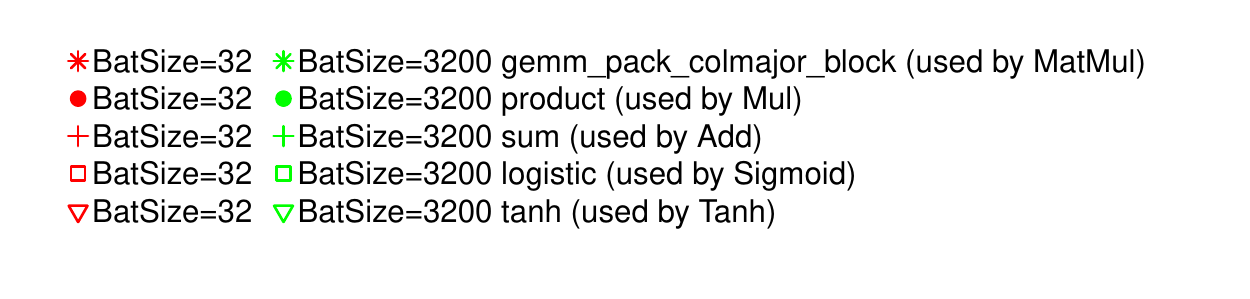}
  \end{center}
    \caption{Roofline chart of the RankNet on the CPU Platform. The chart hightlights the kernels- MatMul, Mul, Add, Sigmoid and Tanh. The position changes from the red dots to green dots, mostly higher Gig OPS values and some with higher AIs, are the reasons why the larger batch\_size had better performance.}
    \label{fig:roofline}
\end{figure}
As a LSTM based model, RankNet contains the kernel operations of matrix multiplication(MatMul), element-wise product(Mul), Add, Sigmoid, and Tanh identified from the architecture of an LSTM cell. 
In the test on CPU, these kernel operations account for over 75\% of the overall walltime, while MatMul alone account for about half.
The roofline chart in Figure \ref{fig:roofline} further explains why large batch\_size training is more efficient.
\begin{figure}[ht]
  \begin{center}
    \includegraphics[width=0.5\linewidth]{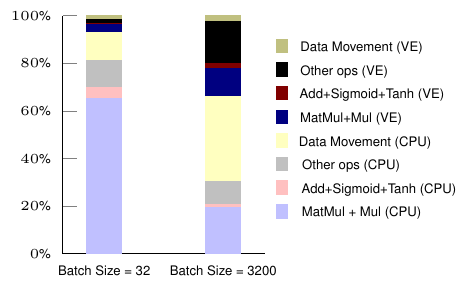}
  \end{center}
    \caption{Operation breakdown for VE/CPU hybrid system, for different batch\_size.}
    \label{fig:gpu-ve-bsize}
\end{figure}
The fastest device for RankNet training depends on the parameter batch\_size and configuration, as shown in Fig.~\ref{fig:impact_BatchSize}. Since offloading to accelerators incurs overheads of extra data movement and function calls, GPU or VE is faster than CPU only when the performance gain from offload can offset the overhead. 
As shown in Figure \ref{fig:gpu-ve-bsize}, when \emph{batch\_size} was set to the default 32, about only 7\% work load were offloaded to the VE, thus the impact on the overall performance comparing to CPU-only result is very minimal. With the \emph{batch\_size} at 3200, there were about 35\% were offloaded to the VE. As VE has better efficiency to do vectorlized and matrix operations, the performance gain from the offload can offset the overhead from data movement and the offloading process.
We test two implementations of LSTM on GPU. The first one is the operation-by-operation implementation similar to what CPU and VE has. The other is CudnnRNN optimized implementation\citet{appleyard2016optimizing, cudnnrnn} which includes combining and streaming matrix multiplication operations to improve the parallelism as well as fusing point-wise operations to reduce data movement overhead and kernel invocation overhead, among other tricks for more complex model.  
GPU run with the operation-by-operation implementation of LSTM shows similar trend as VE does. 
CudnnRNN optimized approach always show the best performance in our experiments. Profiling results with nvprof reveals that the optimizations successfully reduce the number of operation calls, with only 39\% MatMul operations and 1\% scalar(product, sum, logistic, and tanh) left even in small batch\_size=32.  
\section{Related Work}
\textbf{\emph{Forecasting in general:}}
Forecasting is a heavily studied problem interested across domains. 
Classical statistical methods(e.g,ARIMA and exponential smoothing) and machine learning methods(e.g, SVR, Random Forest and XGB) are wildly applied in the problems with few of randomness, non-stationary and irregularity. RNN-based autoregressive models are successful in many forecasting problems. In this paper, we also adopt the RNN-based model following the general encoder-decoder architecture. While Transformer\citet{vaswani_attention_2017}  shows better performance than RNN in many tasks, we find that it works not as good as RNN-based model in the car racing forecasting problem. \citet{li_enhancing_2019} also discusses the limitations of Transformer.
\emph{decomposition to address uncertainty.}
To deal with the problem of high uncertainty, decomposition and ensemble is often used to separate the uncertainty signals from the normal patterns and model them in dependently. \citet{qiu_empirical_2017} utilizes the Empirical Mode Decomposition~\citet{huang_empirical_1998} algorithm to decompose the load demand data into several intrinsic mode functions and one residue, then models each of them separately by a deep belief network, finally forecast by the ensemble of the sub-models. 

Another type of decomposition occurs in local and global modeling. 
ES-RNN \citet{smyl_hybrid_2020}, 
winner of M4 forecasting competition\citet{m4}, 
hybrids exponential smoothing to capture non-stationary trends per series and learn global effects by RNN, ensembles the outputs finally. 

Similar approaches are adopted in DeepState \citet{rangapuram_deep_2018-1}, DeepFactor \citet{wang_deep_2019}. 

In this work, based on the understanding of the cause effects of the problem, we decompose the uncertainty by modeling the causal factors and the target series separately and hybrid the sub-models according to the cause effects relationship. 
Different from the works of counterfactual prediction~\citet{alaa_deep_2017,hartford_deep_2017}, we do not discover causal effects from data.
\emph{modeling extreme events.} 
Extreme events~\citet{kantz_dynamical_2006} are featured with rare occurrence, difficult to model, and their prediction are of a probabilistic nature.  
Autoencoder shows improve results in capturing complex time-series dynamics during extreme events, such as \citet{laptev_time-series_2017} for uber riding forecasting and \citet{yu_deep_2017} which decomposes normal traffic and accidents for traffic forecasting. 
\citet{ding_modeling_2019} proposes to use a memory network with attention to capture the extreme events pattern and a novel loss function based on extreme value theory.
In our work, we classify the extreme events in car racing into different categories, model the more predictable pit stops in normal laps by MLP with probabilistic output. Exploring autoencoder and memory network can be one of our future work.
\emph{express uncertainty in the model.}
\citet{gal_dropout_2016} first proposed to model uncertainty in deep neural network by using dropout as a Bayesian approximation. \citet{yu_deep_2017} followed this idea and successfully applied it to
large-scale time series anomaly detection at Uber.
Our work follows the idea in \citet{salinas_deepar_2019} that parameterizes a fixed distribution with the output of a neural network. \citet{wang_deep_2019-1} adopts the same idea and apply it to weather forecasting.
\textbf{Car racing forecasting:}
\emph{Simulation-based method:}Racing simulation is widely used in motorsports analysis  \citet{heilmeier_race_2018, bekker_planning_2009,Phillips_Simulator_2014}. To calculate the final race time for all the cars accurately, a racing simulator models different factors that impact lap time during the race, such as car interactions, tire degradation, fuel consumption, pit stop, etc., via equations with fine-tuned parameters. Specific domain knowledge is necessary to build successful simulation models. 
\citet{heilmeier_race_2018} presents a simulator that reduces the race time calculation error to around one second for Formula 1 2017 Abu Dhabi Grand Prix. But, the author mentioned that the user is required to provide the pit stop information for every driver as an input. 
\emph{Machine learning-based method:}
\citet{tulabandhula_interactions_2014,choo_real-time_2015} is a series of work forecasting the decision-to-decision loss in rank position for each racer in NASCAR. \citet{tulabandhula_interactions_2014} describes how they leveraged expert knowledge of the domain to produce a real-time decision system for tire changes within a NASCAR race. 
They chose to model the change in rank position and avoid predicting the rank position directly since it is complicated due to its dependency on the timing of other racers' pit stops. In our work, we aim to build forecasting that relies less on domain knowledge and investigate the pit stop modeling.
\section{Conclusion}
In this work, we explore using deep learning models to resolve the challenging problem of modeling sequence data with high uncertainty and extreme events. On the IndyCar car racing data, we find that the decomposition of modeling according to the cause and effect relationship is critical to improving the rank position forecasting performance. 
A combination of encoder-decoder network and separate MLP network that capable of delivering probabilistic forecasting are proposed to model the pit stop events and rank position in car racing. 
In this way, the deep learning modeling approach demonstrates its advantages of automatically features representation learning and modeling complex relationship among multiple time series, reducing the dependency on domain experts in other machine learning and simulation approaches. 
Our proposed model achieves significantly better accuracy than baseline models in both the rank position forecasting task and change of the rank position forecasting task, with the advantages of needing less feature engineering efforts and providing probabilistic forecasting that enables racing strategy optimizations via our deep learning based model. 
There are several limitations to this work. Since there are not many related works, the performance evaluation in this paper is still limited. Another major challenge lies in the restriction of the volume of the dataset. Car racing is a one-time event that the observed data are always limited. Applying transfer learning in this problem could be one important direction of future work.
\medskip
\small
\bibliography{prediction}
\bibliographystyle{iclr2021_conference}

\end{document}